# Debiasing Personal Identities in Toxicity Classification

**Chandra Shekar Bikkanur**
csbikkanur@berkeley.edu

**Apik Ashod Zorian**
zorian@berkeley.edu

**Abstract**

As Machine Learning models continue to be relied upon for making automated decisions, the issue of model bias becomes more and more prevalent. In this paper, we approach training a text classification model and optimize on bias minimization by measuring not only the models performance on our dataset as a whole, but also how it performs across different subgroups. This requires measuring performance independently for different demographic subgroups and measuring bias by comparing them to results from the rest of our data. We show how unintended bias can be detected using these metrics and how removing bias from a dataset completely can result in worse results.

## 1 Introduction

In this work, we develop Machine Learning models that classify a piece of text as toxic or not-toxic. We define "Toxicity" (as it is defined in [1]) as "anything that is rude, disrespectful, or unreasonable that would make someone want to leave a conversation." Throughout this paper, we refer to "subgroup bias" when referring to text in our dataset that targets one or more identities in it, such as specific religion, race, or sexual orientation.

We demonstrate that while a model may perform well on a data set as a whole, it can still demonstrate bias at the subgroup level. We train a variety of models, including TFIDF, LSTM, and BERT, on our dataset and observe each model's respective performance on the test set, as well as its performance on data for individual subgroups. Finally, we take our best performing model and remove identity bias from its training dataset, generating an "identity-free" training set. We train a new BERT model on this unbiased dataset, which serves as our baseline, and show that having subgroup bias in a model's training set is actually necessary to make accurate classifications.

## 2 Background

There has been substantial work in how unintended bias should be measured in text classification. The Conversation AI team, a research initiative founded by Jigsaw and Google, built a model to address this very issue. Their initial model suffered from a significant bias, incorrectly learning to associate the names of frequently attacked identities with toxicity. Last year, members of the Jigsaw team released a study where they used nuanced metrics to measure unintended bias in the data set [1]. Their study focused on measuring a distinct aspect of the model's bias: the skewing of output labels based on subgroup-related content from the dataset. In their work, they investigated threshold-agnostic evaluation metrics to check the performance of the model across different identity-specific groups to shed light on the nuance of unintended bias. As is described in [2], some relevant information may be lost when we measure bias using a single metric, as different types of bias could obscure one another. We utilized the Conversation AI team's nuanced metrics in evaluating our own model's performance on subgroup data.

For our models, we used a TFIDF [3], a 2-layer LSTM [4], and the state-of-the-art BERT model [5]. For our LSTM model, we used GloVe embeddings [6], and for our BERT model we used ELMO embeddings [7].

As explained in [8], blindly applying Machine Learning using arbitrary datasets can result in spreading and magnifying a variety of biases present in the original context. By training a BERT model on a subset of our dataset that is totally void of comments targeting any subgroups, we show how poorly a naïve model performs when faced with classifying comments that do contain subgroup bias.

# 3 Methods

## 3.1 Dataset

Our dataset consists of 1.8 million comments collected on Civil Comments, a platform that brought real-world social cues to comments sections via crowd-sourced moderation community management tools [10]. The platform was shut down in 2017, but public comments were made available as an open archive for research.

Each datapoint consists of a *comment_text* column , which contains the comment, and the toxicity label. The target label is a floating point value between 0.0 and 1.0, where 0.0 is not toxic and 1.0 is very toxic. Table 1 shows a few examples of such comment and toxicity score pairs. Note that the more aggressive the tone and content of the comment gets, the higher the toxicity score.

| Comment Text | Toxicity |
|---|---|
| *I think the earth goes through cycles and we're in a warmer cycle* | 0.0 |
| *My thoughts are that people should stop being stupid and ignorant. Climate change is scientifically proven. It isn't a debate.* | 0.29 |
| *They are liberal idiots who are uneducated.* | **0.81** |

**Table 1**: Examples of comment text and respective toxicity level

For our dataset, we converted the Toxicity values to a Boolean, where any value >= 0.5 is considered toxic, and saved as a Boolean value 1, while values below 0.5 were saved as 0. A subset of these comments are labeled with identity attributes, based on the subgroup(s) referenced in the comment. As the comments were labeled by human volunteers, the dataset does contain different comments which contain the exact same text but are labeled for different identities. Some examples of the identities include race (Black, White, Asian, Latino), religion (Christian, Jewish, Muslim), and sexual orientation (heterosexual, homosexual, bisexual, other). These identities are the subgroups that we target when evaluating our model's performance. One initial observation was that of the 1.8 million comments in the dataset, close to 90% were not toxic, the significance of which is discussed in the Results section.

The values for each subgroup column were also floating point values that ranged between 0.0 and 1.0, depending on how explicitly a comment targeted a specific subgroup. To better isolate subgroup-biased data, we added a new column *identity_bias* which summed all of these subgroup columns for a given example. In our final experimentations, we trained our baseline model only on data vales with *identity_bias* < 0.25. Figure 2 shows a word cloud of the most prevalent words in the subset of data that had an *identity_bias* > 0.75.

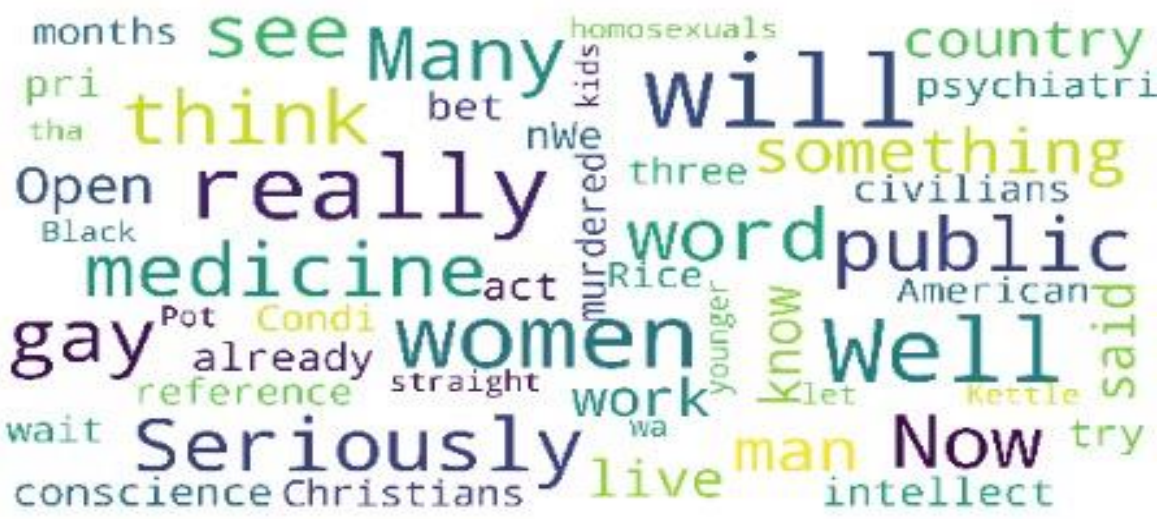

**Figure 1**: Prevalent words from comments with *identity bias* > 0.75

## 3.2 Evaluation Metrics

To evaluate our model's performance, we used metrics that were built on Area Under the Receiver Operating Characteristic Curve (AUC). An advantage of AUC is it is robust to datasets that may have unequal numbers of positive and negative examples. AUC is also a threshold agnostic metric, meaning it is possible to perfectly separate classes if, for example, we have an AUC score of 1.0.
Standard metrics for measuring unintended bias are contingent upon subgroups within the dataset. However, we do not rely solely on calculating AUC on each subgroup to measure our model's performance. Rather, we compare the subgroup data to the rest of the data that does not include the subgroup, which is called the "background" data. Thus, given a subgroup, we can divide our data into negative subgroup examples, positive subgroup examples, negative background examples, and positive background examples. (Note: positive examples refer to toxic comments, while negative examples refer to non-toxic). We use three AUC-based bias metrics defined in Nuanced Metrics for Measuring Unintended Bias[1] paper:

1. **Subgroup AUC**: The AUC exclusively on examples from a given subgroup. This represents the model's ability to separate and accurately classify positive and negative examples from a subgroup

2. **BPSN AUC** (Background Positive, Subgroup Negative): AUC on negative subgroup examples and positive background examples. If this value is low, it means that the model has trouble separating non-toxic subgroup examples from toxic background examples. A low score would mean that the model will predict many false positive, giving high toxicity scores for non-toxic subgroup examples

3. **BNSP AUC** (Background Negative, Subgroup Positive): AUC on negative background examples and positive subgroup examples. If this value is low, it means that the model has trouble separating toxic subgroup examples from non-toxic background examples. A low score would mean the model will predict many false negative, giving low toxicity scores for toxic subgroup examples

While [1] focused on showing the importance of these metrics in testing unintended bias, our goal is to take this a step further by calculating these metrics on models that would be trained on a dataset with comments that contained no subgroup-bias, and comparing our results with a model whose training set did contain this bias.

## 4 Models

Table 3 shows the results from our experimentation with different types of NLP models. All three models were trained on the same training (1.6M samples) and testing (300K) data sets. For our TFIDF, we used a TFIDF Vectorizer and logistic regressor. For our LSTM, we had one embedding layer, one LSTM layer with 0.1 dropout and 50 sized input, and two dense layers, one with 50 sized input and relu activation, and another with 2 sized input and softmax activation. We fit a tokenizer with a max number of words set to 10,000 and padded our text with a max sequence length of 220 to ensure our comments would all be of equal length.

For BERT, we used a pre-trained base uncased model. The model's architecture consists of ElMo embeddings, 12 layers, 768 hidden states, 12 heads and 110M parameters. We used the *bertForSequenceClassification* class of the pre-trained BERT model to fine tune the parameters for text classification. Hyper parameters for the model are bert_adam optimizer, batch size of 32 and learning rate of 0.00002.

Results of our models can be seen in Table 2. While the Overall AUC was relatively close for LSTM and BERT, when observing the Subgroup AUCs, we could see that BERT outperformed for every identity. The steady improvement of AUCs from least complex model (TFIDF) to most complex (BERT) was expected. It was interesting to note that ranking of subgroup AUC scores was nearly identical across models, with Female, Male, and Christian scoring as 3 of the 4 highest in every case. This is most likely due to the fact that these three subgroups were represented much more in the training set than the rest of the subgroups.

### 4.1 Baseline

For our baseline model, we trained a model on identity-neutral data, which we will call our Naive Model. Our goal was to compare its results to a model trained on an equal-sized dataset of both identity-neutral and identity-targeted data, which we will call our Mixed Model. To train our Naive Model, we used only data that had *identity_targeted* values < 0.25, while our Mixed Model was trained on an equal-sized dataset but without filtering subgroup targeting. As BERT had clearly outperformed our other models, we initialized a pretrained BERT model similar to our previous model, and trained it on 360975 samples of unbiased data. We then did the same experiment using an equal amount of mixed data. Table 2 shows the distribution on the datasets for the two models.

| | Mixed Model | Naive Model |
|---|---|---|
| Total Number of Comments | 360975 | 360975 |
| Non-Subgroup Toxic Comments | 82933 | 91451 |
| Non-Subgroup Not Toxic Comments | 1080902 | 1198776 |
| Subgroup Toxic Comments | 20308 | N/A |
| Subgroup Not Toxic Comments | 106084 | N/A |
| Not Toxic/Toxic Comments Ratio | 5 to 1 | 13 to 1 |

**Table 2**. Distribution of training data for Mixed and Naïve model

| Model | Overall AUC | Subgroup AUC | | | | | | | | |
|---|---|---|---|---|---|---|---|---|---|---|
| | | **Homo-sexual** | **Black** | **White** | **Muslim** | **Jewish** | **Female** | **Mental Illness** | **Male** | **Christian** |
| TFIDF | 0.740 | 0.669 | 0.674 | 0.693 | 0.670 | 0.657 | 0.713 | 0.734 | 0.730 | 0.707 |
| LSTM | 0.927 | 0.783 | 0.787 | 0.800 | 0.826 | 0.845 | 0.868 | 0.869 | 0.870 | 0.888 |
| BERT | 0.930 | **0.848** | **0.842** | **0.855** | **0.873** | **0.894** | **0.957** | **0.926** | **0.926** | **0.929** |

**Table 3**: Overall and Subgroup AUC scores for TFIDF, LSTM, and BERT models

| | Mixed Model | | | Naive Model | | | |
|---|---|---|---|---|---|---|---|
| **Subgroup** | **Subgroup AUC** | **BPSN AUC** | **BNSP AUC** | **Subgroup AUC** | **BPSN AUC** | **BNSP AUC** | **Subgroup Size** |
| homosexual | 0.861 | 0.87 | 0.973 | 0.817 | 0.944 | 0.927 | 2223 |
| black | 0.866 | 0.85 | **0.977** | 0.837 | 0.913 | 0.955 | 2959 |
| white | 0.870 | 0.87 | 0.974 | 0.85 | 0.918 | 0.958 | 5003 |
| muslim | 0.870 | 0.88 | 0.9745 | 0.87 | 0.953 | 0.938 | 4229 |
| jewish | 0.891 | 0.91 | 0.96 | 0.891 | 0.952 | 0.943 | 1529 |
| psych | 0.927 | 0.92 | 0.97 | 0.918 | 0.920 | **0.973** | 978 |
| male | 0.937 | 0.94 | 0.969 | 0.924 | 0.949 | 0.963 | 8894 |
| female | 0.937 | **0.955** | 0.959 | 0.927 | 0.954 | 0.960 | 10690 |
| christian | **0.938** | 0.936 | 0.973 | **0.927** | **0.973** | 0.935 | 8285 |

**Table 4**: Subgroup, BPSN, and BNSP AUC scores for Mixed and Naïve Models

## 5 Results

Table 4 shows the results of our experiment with our Mixed and Naïve models. Both models scored almost identical total AUCs, with our Naïve model scoring .973 and our Mixed model scoring 0.970. However, when we observe the Subgroup AUCs displayed in Figure 4b, we can see that our Mixed Model performed better on every subgroup. Since our Mixed Model was exposed to comments that targeted specific subgroups, it was able to more clearly separate toxic data from non-toxic examples at the subgroup level when compared to the Naïve Model. Figure 2 contrasts the models' scoring distribution across different subgroups. We can see that the Mixed Model's does a much better job of classifying the toxic examples, while the Naïve Model often has toxic examples clustered around scores that should obviously be not-toxic.

Observing the BPSN and BNSP AUCs yielded some interesting findings. For almost every subgroup, the Mixed model has a higher BNSP AUC, which would imply that the Naive model is more susceptible to *false negatives* in these subgroups. This makes sense, as it has not been exposed to identity-targeting comments. Furthermore, while we had seen that our data was skewed towards non-toxic comments, this was even more true when we observed the training data for our Naive Model (Table 2). After extracting unbiased data using the *subgroup_targeted* variable, we calculated 13 toxic comments for every 1 toxic comment in our unbiased training data, as opposed to 5 to 1 ratio for our Mixed model. This also explains why our Naive model may default to non-toxic predictions on subgroup examples in the test set.

The opposite is true for BPSN, as our Naive model scored a higher BPSN AUC for nearly every subgroup. This implies that our Mixed model is more susceptible to *false positives* when classifying subgroup specific comments. We were initially surprised to see these results, as we imagined a model that was fully exposed to subgroup targeting data would have been better at classifying subgroup comments. Per Table 2, the number of positive non-identity comments were roughly the same for both models' training datasets. The Naive Model does, however, have an advantage (albeit, a cheap advantage) when classifying negative subgroup data. As it has not been exposed to any subgroup data, it has been trained on an overwhelmingly large number of non-toxic comments. Therefore, if the model struggles when classifying a subgroup targeting comment, it may default to classifying it as non-toxic. Doing this enough will lead to better separability between Subgroup Negative and Background Positive comments.

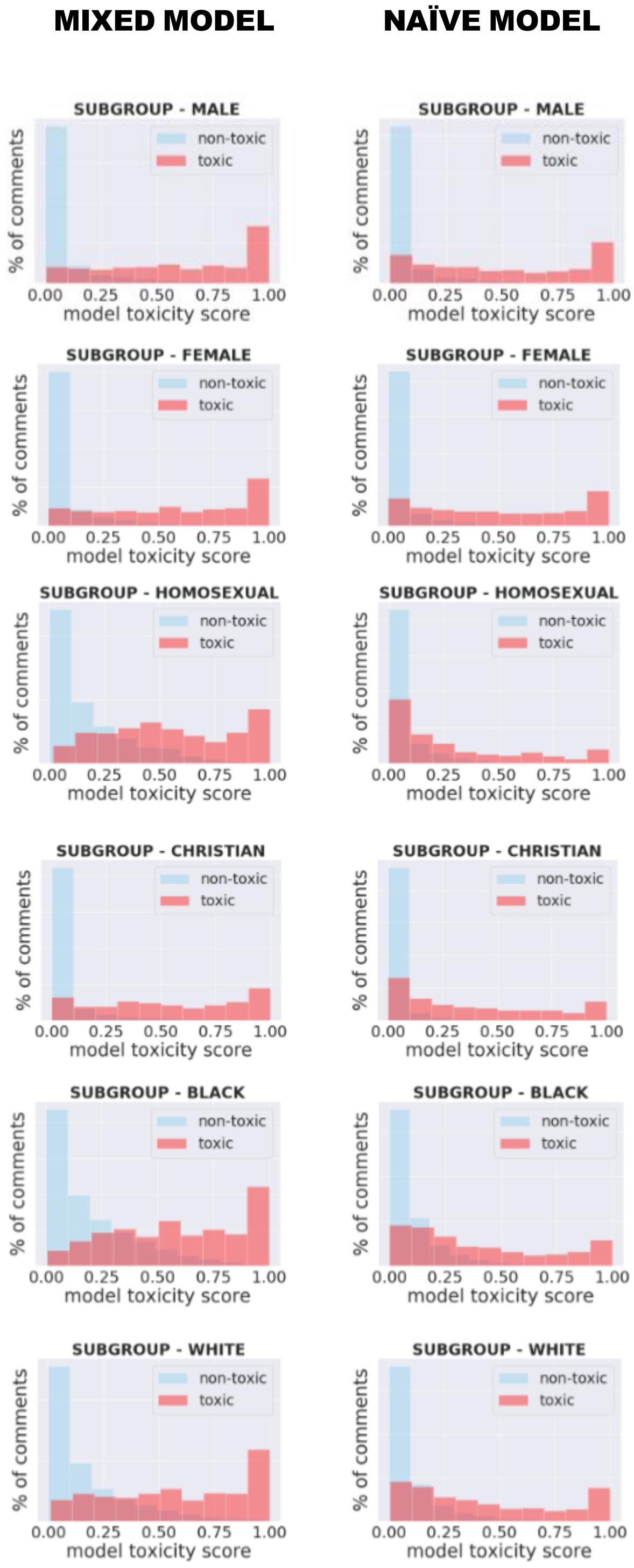

**Figure 2**. Subgroup scoring distribution for Mixed (left) and Naïve (right) models

## 6 Conclusion

This project demonstrated the importance of having bias in the data used to train a text classification model. We showed that a model that was not exposed to identity-targeting text would, on the surface, perform just as well, yet when we dive into examples targeting specific subgroups, we can see that a biased model will outperform it. And while our data was somewhat skewed towards non-toxic comments, our model was still robust enough to be able to score an AUC of over .93 in both cases. We also showed the power of the BERT model and how accurately it was able to classify comments in our data set. The fact that the model was trained solely on text, with no access to any of the subgroup features, and was able to score so well on subgroup data is a testament to the sheer strength and cogency of today's state-of-the-art NLP models. Finally, we showed how nuanced metrics for testing bias in ML models can help uncover characteristics in models that are not as apparent on the surface.

Future iterations of this project could involve digging into the structure of BERT, fine tuning the model by using experimenting with different hyperparameters, adding more layers, or potentially using an ensemble flow with multiple models. We could also experiment with other datasets and add annotations for other subgroups, such as political affiliations and biases to different states in the U.S., as much of our data had electoral information.

Another factor we noticed in our dataset that would need to be addressed is the element of sarcasm. In [9], the authors describe how sarcasm can flip the polarity of a positive sentence and negatively affect polarity detection. We noticed plenty of this in our dataset and acknowledged that models will suffer in classifying comments as toxic if they are unable to discern the irony in a comment. Finally, debiasing word embeddings could also be an improvement worth tackling. We have seen some gender biases in GloVe embeddings, such as "Man is to Doctor as Woman is to Nurse". While we were able to use cosine similarities to identify a few of these, neutralize them, regenerate embeddings, and retrain our model, we did not see enough improvement in our results to mention this. However, more can be done in looking at false positives and false negatives in a models prediction, identifying the cause of these mistakes, and seeing if neutralizing a biased word association in the embeddings would result in improved performance.